\documentclass{article}

\usepackage[preprint]{neurips_2024}

\usepackage{microtype}
\usepackage{graphicx}
\usepackage{subfigure}
\usepackage{booktabs} % for professional tables
\usepackage{float}
\usepackage{caption}

\usepackage{hyperref}
\usepackage{amsmath}
\usepackage{ulem}

\usepackage[ruled,vlined]{algorithm2e}
\usepackage[noend]{algpseudocode}

\usepackage{amsmath}
\usepackage{amssymb}
\usepackage{mathtools}

\usepackage[capitalize,noabbrev]{cleveref}

\newcommand\review[1] {{\color{black}{#1}}}

\usepackage[disable,textsize=tiny]{todonotes}
\usepackage[utf8]{inputenc} % allow utf-8 input
\usepackage[T1]{fontenc}    % use 8-bit T1 fonts
\usepackage{hyperref}       % hyperlinks
\usepackage{url}            % simple URL typesetting
\usepackage{booktabs}       % professional-quality tables
\usepackage{amsfonts}       % blackboard math symbols
\usepackage{nicefrac}       % compact symbols for 1/2, etc.
\usepackage{microtype}      % microtypography
\usepackage{xcolor}         % colors

\title{Sampling-based Distributed Training with Message Passing Neural Network}

\author{%
  Priyesh Kakka\\[0.2cm]
  Department of Aerospace and Mechanical Engineering\\
  University of Notre Dame, USA\\
  \texttt{pkakka@nd.edu}\\[0.5cm]
  \And
  Sheel Nidhan\\[0.2cm]
  Office of CTO\\
  Ansys, Inc., USA\\
  San Jose, CA 95134\\
  \texttt{sheel.nidhan@ansys.com}\\[0.5cm]
  \And
  Rishikesh Ranade\\[0.2cm]
  Office of CTO\\
  Ansys, Inc., USA\\
  Canonsburg, PA 15317\\
  \texttt{rishikesh.ranade@ansys.com}\\[0.5cm]
  \And
  Jay Pathak\\[0.2cm]
  Office of CTO\\
  Ansys, Inc., USA\\
  Canonsburg, PA 15317\\
  \texttt{jay.pathak@ansys.com}\\[0.5cm]
  \And
  Jonathan F. MacArt\\[0.2cm]
  Department of Aerospace and Mechanical Engineering\\
  University of Notre Dame, USA\\
  \texttt{jmacart@nd.edu}%
}

\begin{document}

\maketitle
\begin{abstract}
In this study, we introduce a domain-decomposition-based distributed training and inference approach for message-passing neural networks (MPNN). Our objective is to address the challenge of scaling edge-based graph neural networks as the number of nodes increases. Through our distributed training approach, coupled with Nyström-approximation sampling techniques, we present a scalable graph neural network, referred to as DS-MPNN (D and S standing for distributed and sampled, respectively), capable of scaling up to $O(10^5)$ nodes. We validate our sampling and distributed training approach on three cases: (a) a Darcy flow dataset (b) steady RANS simulations of 2-D airfoils and (c) steady RANS simulations of 3-D step flow, providing comparisons with both single-GPU implementation and node-based graph convolution networks (GCNs). The DS-MPNN model demonstrates comparable accuracy to single-GPU implementation, can accommodate a significantly larger number of nodes compared to the single-GPU variant (S-MPNN), and significantly outperforms the node-based GCN.
\end{abstract}

\section{Introduction and Related Work}\label{sec:introduction}
The application of machine learning algorithms to build surrogates for partial differential equations (PDEs) has seen a significant push in recent years \cite{li2020fourier, raissi2019physics,lu2021learning, pfaff2020learning}. While the majority of these ML models for PDEs surrogate modeling are CNN-based~\citep{zhu2019physics, ranade2022composable,ren2022phycrnet}, the introduction of graph neural networks \citep{kipf2016semi} has made graph-based modeling increasingly common. Traditional numerical methods for real-world applications predominantly rely on mesh-based unstructured representations of the computational domains \cite{mavriplis1997unstructured} owing to the ease of (a) discretization of complex domains and (b) extension to adaptive mesh refinement strategies to capture multi-scale physical phenomenon like turbulence, boundary layers in fluid dynamics, etc. These mesh-based unstructured formulations can be seamlessly formulated on graphs. Furthermore, a graph-based representation eliminates the need for interpolating data onto a structured Euclidean grid -- a step inherent in CNNs, further reducing cost and error. In recent years, these graph-based models have also been applied to crucial domains like weather modeling~\citep{lam2022graphcast}, biomedical science~\citep{fout2017protein}, computational chemistry~\citep{gilmer2017neural, duvenaud2015convolutional}, etc. 

From the perspective of physical system dynamics modeling, we can divide the GNN-based methods into two broad categories: (a) node-based approaches and (b) edge-based approaches. The node-based approaches, including graph convolution networks (GCNs) ~\citep{kipf2016semi}, GraphSAGE ~\citep{hamilton2017inductive}, and graph attention networks (GATs) ~\citep{velivckovic2018graph}, are easier to scale with an increasing number of elements in an unstructured grid setting. However, they lack the mechanism of message-passing among nodes which is crucial to the modeling of spatial relationships between the nodes in complex PDEs. Hence, the edge-based graph methods dominate the paradigm of graph-based modeling for physical systems, such as Graph Network-based Simulators~\citep{sanchez2020learning}, and its evolution MeshGraphNets (MGN)~\citep{pfaff2020learning}. ~\citet{pfaff2020learning} indicate that these edge-based graph methods outperform node-based GCNs in accuracy and stability, and can be applied to a wide variety of simulations. Another approach to edge-based methods for PDE modeling is message-passing neural networks~\citep{gilmer2017neural, li2020neural} where the input and the output solution space are invariant to mesh grids and independent of discretization~\citep{li2020neural}. Message-passing neural networks (MPNN)~\citep{simonovsky2017dynamic} condition weights around the vertices based on edge attributes and have properties to learn the PDE characteristics from a sparsely sampled field.

While edge-based methods are accurate in modeling physical systems, their memory requirement scales with the number of edges in a mesh. Hence these methods do not scale for modeling realistic physical problems where the number of nodes can be substantially higher ($O(10^5-10^6)$). ~\citet{bonnet2022an} attempted to solve this issue by randomly sampling a set of nodes and edges based on Nystr\"om approximation~\citep{li2020neural} and memory constraints of the GPUs available, thus creating a sparse domain. The sampling method yields similar errors as modeling complete graphs. Despite many attempts, edge-based methods, especially MPNNs, remain memory-intensive, constrained by the current limits of single GPU memory. \citet{stronisch2023multi} proposed an approach implementing multi-GPU utilization for MeshGraphNets~\citep{pfaff2020learning}. In their methodology, the domain is partitioned across various GPUs, enabling inter-node communication. However, this communication is limited to node-based features within the latent space, excluding interaction among edge-based features. The errors observed in the multi-GPU setup proposed by \cite{stronisch2023multi} were notably higher compared to their single GPU training. This highlights the imperative need to devise methodologies for training edge-based graph models on multi-GPU setups without compromising accuracy.

We present a sampling-based distributed MPNN (DS-MPNN) that involves partitioning the computational domain (or graph) across multiple GPUs, facilitating the scalability of edge-based MPNN to a large number of nodes. Here, `distributed' implies partitioned spatial domains that are put on different GPUs which can be on the same or different machines. This scalability holds significant practical value for the scientific community. We combine traditional domain decomposition techniques used in engineering simulation and message-passing among GPUs to enable the training of message-passing neural networks for physical systems with no or minimal loss in accuracy.

Our two key contributions are:
\begin{enumerate}
    \item We devise a method inspired by domain-decomposition based parallelization techniques for training and inference of MPNN on multiple GPUs with no or minimal loss in accuracy.
    \item We demonstrate the scaling and acceleration of MPNN training for graphs with DS-MPNN to $O(10^5)$ nodes through the combination of multi-GPU parallelization and node-sampling techniques
\end{enumerate}

\section{Model Description}\label{sec:model}
The message-passing graph model used in this work uses a convolutional graph neural network with edge conditioning~\citep{simonovsky2017dynamic} together with random sampling \citep{li2020neural,bonnet2022an}.

\subsection{Graph construction}\label{sec:graph_construction}
\begin{figure}
\centering
\includegraphics[width=\linewidth]{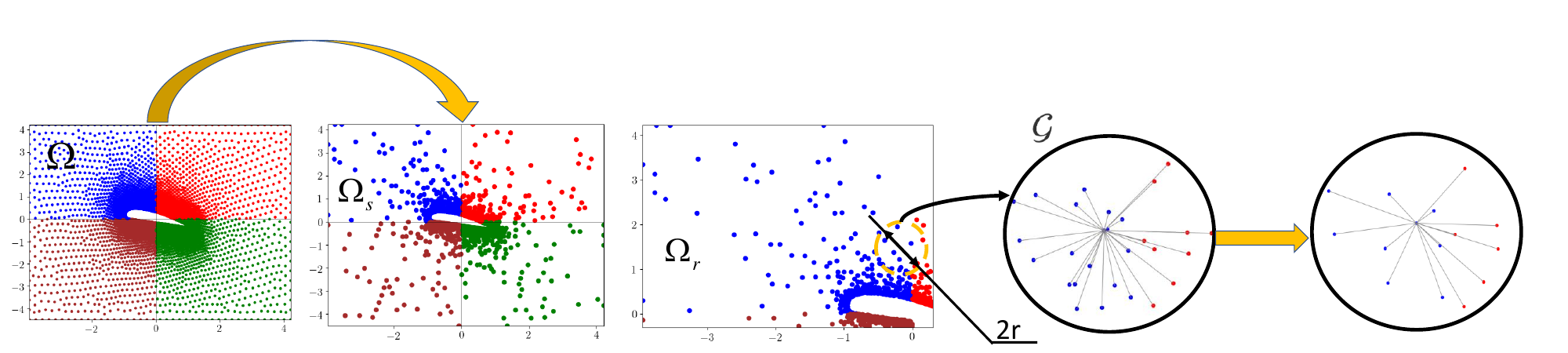}
\caption{Graph kernel $\mathcal{G}$ construction for an individual node in the low-fidelity AirfRANS dataset. Yellow arrows indicate node and edge sampling; various colors denote distinct computational domains on separate GPUs; and $\Omega_{r}$ represents the distributed domain, encompassing an overlap from neighboring domains with a length of kernel radius $r$.}
\label{fig:graph construction}
\end{figure}

Consider a set of nodes \(V \in \mathbb{R}^d\) on the domain \(\Omega \in \mathbb{R}^d\). A subset \(\Omega_s \subset \Omega\) is formed by randomly sampling nodes, with \(|V_s| = s\). Graph kernels \(\mathcal{G}_i = (V_{i}, \mathcal{E}_i)\) are constructed using these sampled nodes $v_i \in V_s$ as centers within a radius \(\rho\). Edge connections \(e_{ij} \in \mathcal{E}_i\) are established between node $j$ within a specified radius and the respective central nodes $i$. For edges $|\mathcal{E}_i| > n_e$, $n_e$ edges are further randomly sampled from $|\mathcal{E}_i|$. The representation of this graph kernel construction is shown in figure~\ref{fig:graph construction}. Each node \(v_i\) and the connecting edges \(e_{ij}\) are assigned attributes or labels, $v_i^{l}$ and $e_{ij}^{l}$, respectively.  In PDE modeling, the nodal and edge attributes can represent the initial functional space \(\mathcal{F}_{inp}\) or the initial conditions. Edge attributes $e_{ij}^{l}$ in this work are derived by calculating the relative difference between node coordinates and attributes $(v^{l})$ of nodes $i$ and $j$. 

At the core of this model is the concept of using edge-conditioned convolution~\citep{simonovsky2017dynamic} to calculate the node attribute $v_{i}^{l}$ in one message-passing step by summing the product of weights $\mathcal{K}_{\phi}$, based on the edge attributes $e_{ij}$ and neighboring node attributes $v_{j}$ in $\mathcal{G}_i$, giving
%\vspace{-0.2cm}
\begin{equation}
{v_{i}}^{l} = \frac{1}{|\mathcal{E}_i|}\sum_{j=1}^{j \le n_{e}}{\mathcal{K}_{\phi}({e^{l-1}_{ij}};{\theta}}){v_{j}}^{l-1} + b.
\end{equation}
Here, $\theta, b$ are neural network parameters.  $l$ corresponds to the message-passing step among `radius hops' $h$.  $e_{ij}^{l}$ is updated based on new values of $v_i^{l}$. Edge conditioning in the convolutional process renders the model adept at handling non-uniform grid points or graph structures, typical in physics simulations involving boundary layers and shocks, where grid density varies significantly across the field.

\subsection{Model algorithm} \label{sec:model_algorithm}
The current model, adapted from~\cite{bonnet2022an}, is composed of an encoder ($\mathcal{N}_{e}$), decoder ($\mathcal {N}_{d}$) and message-passing network ($\mathcal{K}_{\phi}$). This encoder transforms the initial node attributes, \(v_{i}^{l={0}}\), where $l={0}$ indicates the initial or lower-level state of these attributes, into a latent representation. The latent space representation of node attributes at the \(l\)-th message-passing step is represented as \(v_{L,i}^{l}\), with $L$ signifying the latent representation of the node attribute $v_i^{l}$. The transformation process is governed by the following equations:

\begin{align*}
    v_{L,i}^{l=0} &= \mathcal{N}_{e}(v_{i}^{l={0}},\theta) \tag{i},\\
    v_{L,i}^{l+1} &= \frac{1}{|\mathcal{E}_i|} \sum \mathcal{K}_{\phi}(e^{l}_{ij};\theta) v_{L,j}^{l} + b \tag{ii},\\
% \end{align*}
% \begin{align*}
    v_{i}^{l+1} &= \mathcal{N}_{d}(v_{L,i}^{l+1},\theta) \tag{iii},\\
    e_{ij}^{(l+1)} &\leftarrow v_{i}^{l+1} - v_{j}^{l+1} \tag{iv}.
\end{align*}
The process iterates through equations (ii) to (iv) for a total of $l= 0, h-1$ message-passing steps, representing \(h\) radius hops in the graph, to update the node and edge attributes. The objective of the neural network denoted as \(\mathcal{N}(\mathcal{N}_e, \mathcal{N}_d, \mathcal{K}_\phi)\), is to learn the mapping from the initial function space, \(\mathcal{F}_{\text{inp}}\), to the final function space $\mathcal{F}_\text{out}$, giving:

\begin{equation}
\mathcal{N}: \mathcal{F}_{\text{inp}} \rightarrow \mathcal{F}_\text{out}.
\end{equation}

This framework facilitates the application of graph neural networks in solving PDEs by iteratively updating and transforming node and edge attributes within the graph structure.
\section{Methodology}\label{sec:methodology}
We now turn to DS-MPNN, our framework devised to train edge-based graphs on many GPU systems. DS-MPNN incorporates communication strategies tailored for MPNN-based graph methodologies across multiple GPUs. In this framework, prior to the formulation of a graph kernel \(\mathcal{G} = (V_s, \mathcal{E})\), the computational domain \(\Omega\) is partitioned into distinct subdomains and each subdomain is allocated to a GPU. Each GPU is allocated a distinct subdomain \(\Omega_{r} \subset \Omega\). This arrangement entails dividing \(\Omega\) into \(n_{\text{procs}}\) (total number of GPUs available for parallelization) subdomains \(\Omega_{r}\), with each subdomain featuring an extended overlap of length \(l\). In all our runs, unless explicitly mentioned, we set $l=r$ to ensure that nodes at the edges of any given spatial partition have complete kernels This extended overlap is pivotal for ensuring comprehensive kernel construction at the interior edges of the domain, thereby circumventing the issues of incomplete kernel formation that can result in discontinuities in the predicted solutions across the subdomain boundaries. This approach is depicted in figure \ref{fig:graph construction}

\begin{figure*}[ht]
\centering
\includegraphics[width=1.08\linewidth]{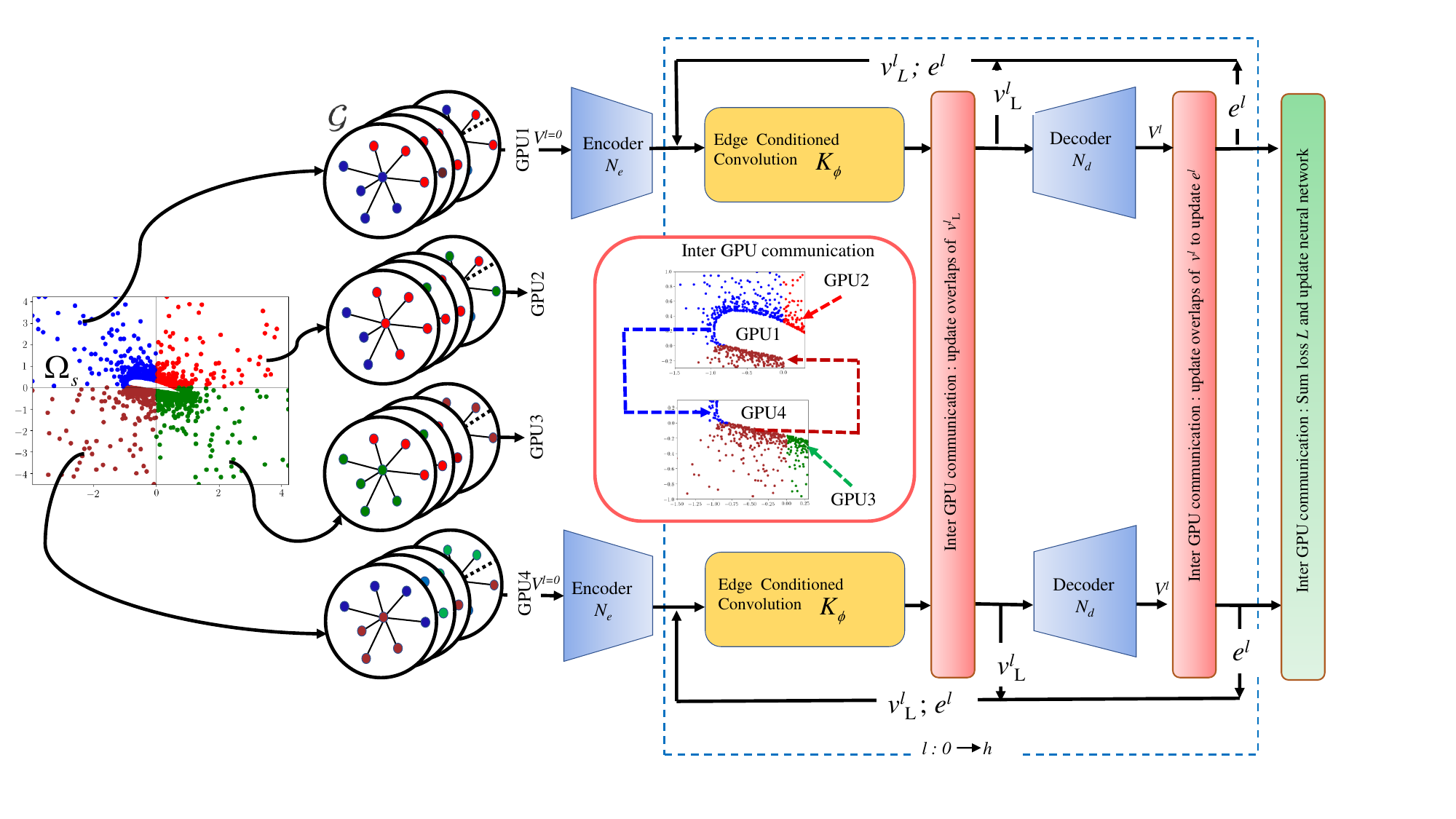}
\caption{Methodology representing graph kernels $\mathcal{G}$ from separate distributed domains of an AirfRANS dataset being processed on four individual GPUs. Inter GPU communication represents the exchange of information between the neighboring domains through overlap regions during each hop. After $h$ radius hops, loss $L$ is calculated on the interior points and aggregated over all GPUs to update the neural network parameters.}  
\label{fig:domain-decomp2}
\end{figure*}

In the framework described in section~\ref{sec:model_algorithm}, each domain, accompanied by its kernels, is allocated to distinct GPUs for a series of $h$ hops. The methodology facilitates inter-GPU communication of the latent space node attributes $v^{l}_{i,L}$ through overlapping regions. As depicted in figure~\ref{fig:domain-decomp2}, the overlap area of a given domain is updated from the neighboring domains' interiors. Concurrently, the decoder's output in the physical space, denoted as $v_{i}^{l}$, updates the edge attributes correspondingly.

The computational graph accumulates gradients during the $h$ radius hops that are computed in relation to the total loss function $\mathcal{L}$ across the entire computational domain. This computation encompasses a summation of the domains' interior points across all GPUs. Following this, an aggregation of the gradients from each GPU is performed, leading to a synchronous update of the neural network parameters across all the GPUs - %\Rishi{I think you need to spell out how this is important in improving accuracy}
, utilizing the aggregated gradient. For interested readers, the detailed DS-MPNN training mechanism is provided in the Appendix.

During testing, the algorithm divides the domains into smaller, randomly selected sub-domains $\Omega_{s}$ along with its overlapping regions. These are sequentially fed into the trained model. They are reassembled in post-processing to form the original output dimension $\Omega \in \mathbb{R}^d$. The inference time of these surrogate models is orders of magnitude smaller than the time required to solve the PDEs, as we will show in the subsequent sections.

We investigate a range of PDEs for both structured and unstructured meshes, incorporating a diverse array of mesh sizes. The core of our study involves a detailed comparative analysis of two distinct computational implementations: a single-GPU implementation S-MPNN and the multi-GPU DS-MPNN. The multi-GPU implementation leverages up to 4 GPUs i.e., DS-MPNN4. This scalability is not limited to the four-GPU configuration tested; our methodology is designed to be flexible with respect to GPU count, allowing for expansion beyond the tested range. Such adaptability is essential for handling a variety of computational demands and hardware configurations.

In the multi-GPU setup, domains are equally partitioned based on their coordinates, and a distributed communication package from PyTorch facilitates inter-node communication \citep{paszke2019pytorch}. The neural network architecture employed in this study consists of a 3-layer encoder and a 3-layer decoder, augmented by a convolution kernel. The network, with approximately $700$k parameters, is utilized consistently across all experiments involving DS-MPNN and the baseline Graph Convolution Networks (GCN) This uniformity ensures parameter consistency across different models, and all models are implemented using PyTorch geometric package~\citep{fey2019fast}. GCN here uses 6 hidden layers with a size of 378. Post-encoding, the latent space representation for all experiments apart from parametric studies maintains a consistent dimensionality of 32 attributes. Optimization is conducted using the Adam optimizer, complemented by the OneCycleLR scheduler \citep{smith2019super}, to enhance learning efficiency and effectiveness. 

\section{Experiments}\label{sec:experiments}
\subsection{Darcy Flow (Structured Data)}\label{sec:darcy_flow}

We explore an example utilizing a structured grid, specifically focusing on 2-D Darcy flow that describes the fluid flow through a porous medium. A detailed description of Darcy flow equations is provided in the Appendix. The experimental setup comprises 1024 training samples, each initialized differently, and 30 test samples. Each sample is structured on a grid of \(421 \times 421\), with sampled nodes \(|V_s| = 421\). The attributes of each node consist of two-dimensional grid coordinates ($x_i$ and $y_i$) and the field values of \(a_i\), resulting in node and edge attributes with dimensions \(\mathbb{R}^3\). The graph kernel employed has a radius of 0.2, and the maximum number of edges allowed per node is set at \(n_{e} = 64\). The model is trained over 200 epochs, with each domain in the test case being randomly sampled five times. Evaluations were performed using both a single GPU framework and a multi-GPU configuration, wherein the computational domain was partitioned into four equivalent segments. The number of radius hops employed in this context, denoted by $h$, was set to eight. The test error was quantified using root mean squared error (RMSE), as defined in equation~\ref{eqn: Darcy_RMSE}:

\begin{equation}
\label{eqn: Darcy_RMSE}
\text{RMSE} = \sqrt{\frac{1}{n} \sum_{i=1}^{n} (y_i - \hat{y}_i)^2}
\end{equation}

%%%%%%%%%%% table 1 %%%%%%%%%%%%%%%%%%%%
\begin{table}[ht]
\caption{Comparison of DS-MPNN (on 4 GPUs) with single-GPU implementation (S-MPNN) and GCN.}
\label{tab:Darcy_GPU4}
\centering
\begin{tabular}{@{}lcccc@{}}
\toprule
Method  & GCN & S-MPNN & DS-MPNN4 \\
\midrule
Test Error  & 3.3E-4 & 7.44E-6 & 7.60E-6 \\
\bottomrule
\end{tabular}
\end{table}
%%%%%%%%%%% table 1 %%%%%%%%%%%%%%%%%%%%

Table~\ref{tab:Darcy_GPU4} compares the performance of GCN and MPNN models on the Darcy flow dataset, using both single and 4-GPU setups. GCN under-performs relative to MPNN, highlighting the superiority of edge-based methods in PDE surrogate modeling, even for simpler datasets. This discrepancy is largely due to GCN's omission of edge representation, a key feature in node-based models. Furthermore, the DS-MPNN model shows comparable results on both single and 4-GPU configurations. {The minor variance in test error between these two setups might result from the segmentation of the computational graph across different domains.} 

\begin{table}[ht]
\centering
\caption{Test error based on $L_1$ loss demonstrating the impact of varying parameters: hops ($h$), sampling radius ($r$), maximum number of edges ($n_e$), and nodes sampled ($n_s$).}

\label{tab:Combined_Airf_sm_comp}
\begin{tabular}{@{}lccc@{}}
\toprule
Radius hops ($h$) & $h = 2$ & $h = 4$ & $h = 12$ \\
\midrule
Test Error & 0.029 & 0.024 & 0.023 \\
\bottomrule
\toprule
Radius ($r$) & $r = 0.05$ & $r = 0.12$ & $r = 0.2$ \\
\midrule
Test Error & $0.1$ & $0.032$ & $0.024$ \\
\bottomrule
\toprule
Sampled edges $(n_{e})$& $n_{e} = 32$ & $n_{e}= 64$  & $n_{e} = 128$ \\
\midrule
Test Error & $0.029$ & $0.024$ & $0.024$ \\
\bottomrule
\toprule
Sampled nodes ($s$) & $s = 420$ & $s = 840$ & $s = 1260$ \\
\midrule
Test Error & $0.0244$ & $0.0244$ & $0.0235$ \\
\bottomrule
\end{tabular}
\end{table}

For the Darcy flow dataset, similar to the work by ~\citet{sanchez2020learning} and \citet{pfaff2020learning}, we study the effect of changing crucial hyperparameters for a graph-based model on the performance of our model evaluated using $L_1$ loss. These key hyperparameters include  (a) total number of hops (\(h\)), (b) sampling radius (\(r\)), (c) number of edges (\(n_e\)), and (d) number of nodes (\(s\)). Table \ref{tab:Combined_Airf_sm_comp} shows that increasing these four hyperparameters helps improve the model's accuracy, albeit showing saturation in accuracy beyond a certain threshold. This observation is in agreement with previous works on edge-based techniques by \citet{pfaff2020learning} and \citet{sanchez2020learning}. The key point to note here is that for problems with higher complexity, increasing any of these four hyperparameters will result in an increased GPU memory consumption, underlying the need for a parallelized training paradigm for graph-based models like ours. 

\subsection{AirfRANS (Unstructured Data)}\label{sec:aifrans}

The AirfRANS datasets comprise two distinct unstructured sets, each varying in complexity and scale. The first dataset, as described by~\cite{bonnet2022an}, contains approximately 15,000 nodes per sample, while the second, a higher-fidelity version detailed in~\cite{bonnet2022airfrans} under ``AirfRANS,'' has approximately 175,000 nodes per sample. The DS-MPNN model efficiently processes both datasets, under various Reynolds numbers and attack angles, highlighting its versatility in diverse and complex aerodynamic simulations. A detailed description of PDEs describing the AirfRANS dataset is provided in the Appendix.

\subsubsection{Low-fidelity AirfRANS dataset}\label{sec:low_fid}
The low-fidelity dataset includes a diverse array of airfoils subjected to varying angles of attack, ranging between (\(-0.3^{\circ}\),\(0.3^{\circ}\)). This dataset also incorporates a spectrum of Reynolds numbers, specifically from \(10^6\) to \(5 \times 10^6\). For this investigation, the test error is quantified RMSE in equation~\ref{eqn: Darcy_RMSE}, while mean square error is used as the criterion for optimization during training. The current experimental run comprises \(180\) training samples and \(29\) test samples. Each sample in the dataset is characterized by input node attributes, which include grid coordinates, inlet velocity, pressure, and the distance function between the surface and the node, denoted as \(v_{i}^{l={0}} \in \mathbb{R}^6\). The edge attributes, denoted as \(e_{ij}^{(l)} \in \mathbb{R}^{8}\), include velocity and pressure attributes. These attributes are updated after each radius hop, reflecting changes based on the model output \(y_{i} \in \mathbb{R}^4\), which comprises the $x-y$ velocity components, pressure, and turbulent viscosity. For this specific case, the number of nodes sampled is \(s = 1600\) and the number of edges \(n_{e} = 64\). The training process spanned over \(1000\) epochs, with each test dataset being sampled \(10\) times.

\begin{figure}[htbp]
    \centering
    % Begin the first minipage for the first subfigure
    \begin{minipage}{0.495\columnwidth}
        \centering
        \includegraphics[width=0.9\textwidth]{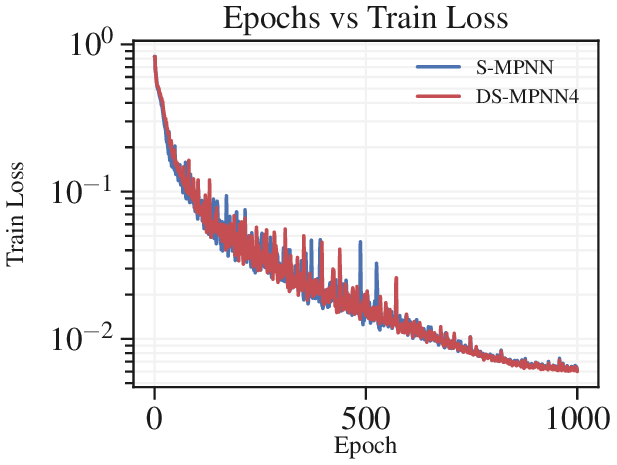}
        % Manual caption for the subfigure
        \textbf{(a)} Scheduler: One cycle learning rate
        \label{fig:cycle_train_loss}
    \end{minipage}
    \hfill % Adds some space between the two minipages
    % Begin the second minipage for the second subfigure
    \begin{minipage}{0.495\columnwidth}
        \centering
        \includegraphics[width=0.9\textwidth]{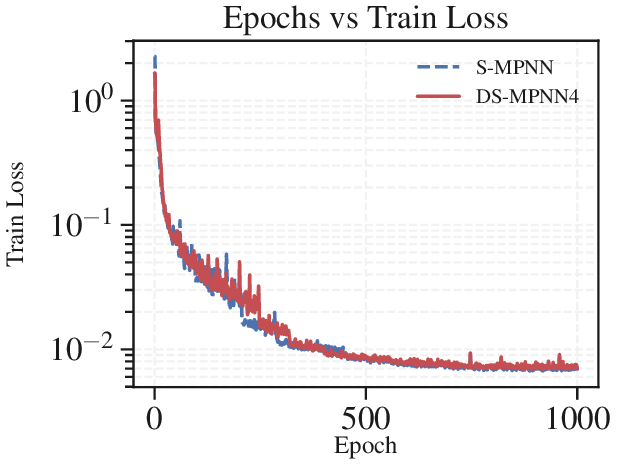}
        % Manual caption for the subfigure
        \textbf{(b)} Scheduler: Dropping learning rate on plateau
        \label{fig:lr_train_loss}
    \end{minipage}
    % Overall caption for the figure
    \caption{Training loss vs. epochs for different schedulers.}
    \label{fig:combined}
\end{figure}

Figure~\ref{fig:combined}(a) illustrates the training loss trajectories of S-MPNN and DS-MPNN4. The convergence patterns, indicated by the overlapping trend in training loss, suggest that DS-MPNN4 achieves a level of training convergence comparable to that of S-MPNN. Furthermore, upon adopting an alternative scheduler strategy—specifically, reducing the learning rate once the training loss plateaus, a similar convergence pattern is observed in figure~\ref{fig:combined}(b). This consistency underscores the training robustness of the DS-MPNN model.

\begin{table}[ht]
\centering
\caption{Comparison of DS-MPNN (on 2 and 4 GPUs) with single-GPU implementation (S-MPNN) and GCN for low-fidelity AirfRANS dataset.}
\label{tab:Airf_sm_comp}
%\resizebox{\columnwidth}{!}{ % Resizes the table to fit within the column
\begin{tabular}{@{}lccccc@{}}
\toprule
Method  & GCN & S-MPNN & DS-MPNN2 & DS-MPNN4   \\
\midrule
RMSE & 0.347 & 0.094 & 0.097 & 0.096 \\
\bottomrule
\end{tabular}
%}
\end{table}

Table~\ref{tab:Airf_sm_comp} shows the test RMSE for GCN, single-GPU MPNN (S-MPNN), and DS-MPNN implemented on 2 and 4 GPUs. The losses are very similar between the single GPU and DS-MPNN runs, underscoring the validity of our algorithm and its effectiveness in handling unstructured grid representation distributed over multiple GPUs. Similar to the Darcy flow results \ref{sec:darcy_flow}, GCN performs worse than MPNNs. However, the accuracy of GCN deteriorates further for the AirfRANS dataset compared to the structured Darcy dataset, highlighting the challenges of using node-based methods for complex PDE modeling. 

\begin{figure}[ht]
\centering
\includegraphics[width=0.8\linewidth]{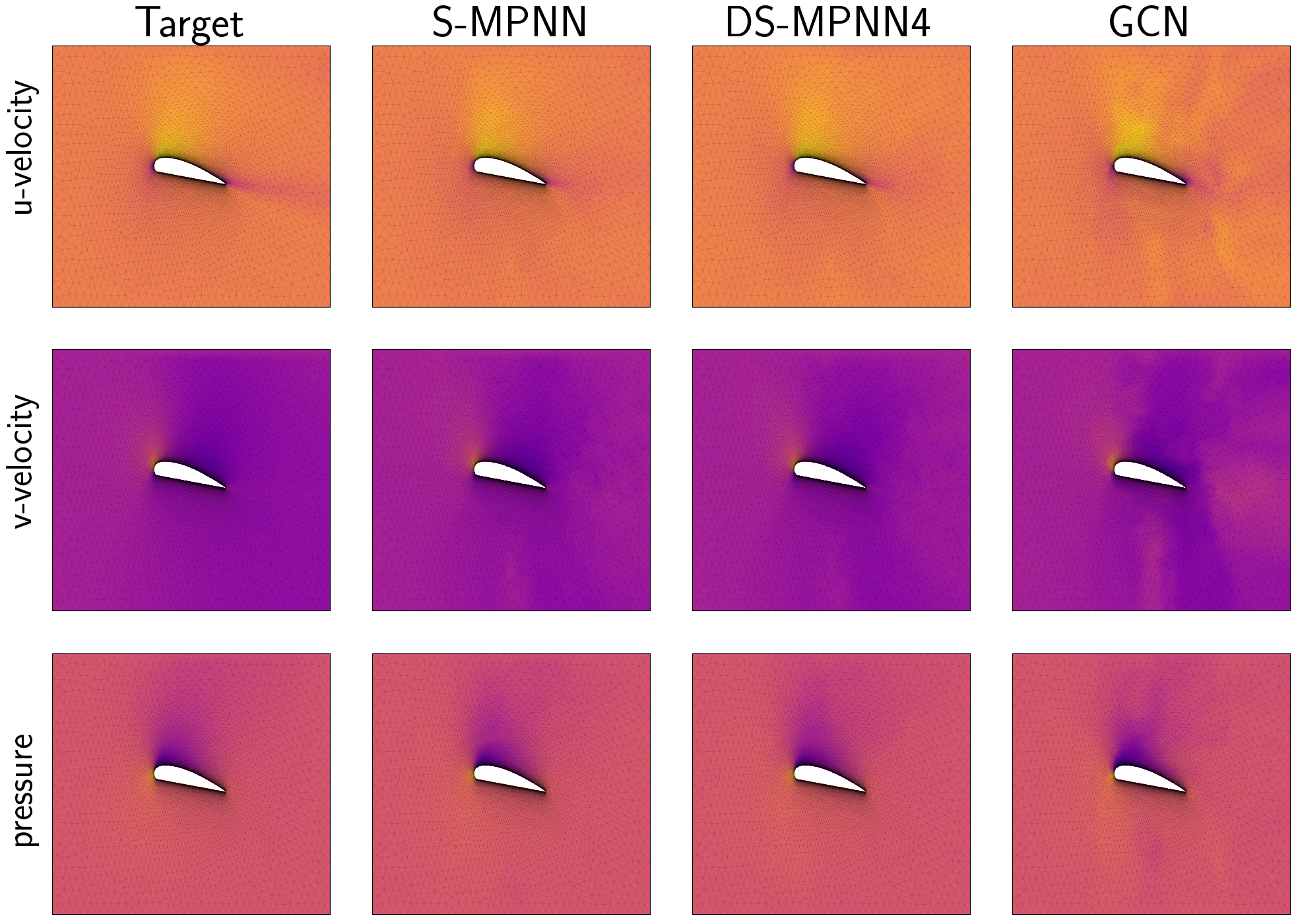}
\caption{Comparison among GCN, S-MPNN, DS-MPNN2 and  DS-MPNN4 for a test sample from low-fidelity AirfRANS dataset.}
\label{fig:airf_sm}
\end{figure}

Figure~\ref{fig:airf_sm} presents a comparative analysis of various models concerning the two velocity components and pressure. It is observed that the performance of the models utilizing a single GPU (S-MPNN) and four GPUs (DS-MPNN4) is comparable. However, the GCN model exhibits inferior predictions, characterized by incorrect flow features around the airfoil across all fields.

\begin{table}[ht]
\centering
\caption{Variation of test RMSE as a function of overlap length ($l$). Kernel radius is set fixed at $r=0.3$.}
\label{tab:Airf_sm_comp_l_variation}
%\resizebox{\columnwidth}{!}{ % Resizes the table to fit within the column
\begin{tabular}{@{}lccccc@{}}
\toprule
  & $l=0$ & $l=0.15$ & $l=0.3$ & $l=0.6$   \\
\midrule
Test RMSE & 0.124 & 0.096 & 0.097 & 0.097 \\
\bottomrule
\end{tabular}
%}
\end{table}

Table 4 presents the test RMSE errors for various extended overlap lengths ($l$) for a fixed kernel radius $r=0.3$. As the overlap radius increases from $0$ to $0.15$, the error decreases significantly. However, beyond $0.15$, the variations are minimal. As previously mentioned in section \ref{sec:methodology}, we set $l=r$ in all our experiments unless explicitly mentioned. This is done to ensure that nodes at the edges of any given spatial partition have complete kernels. It should be noted that the kernel radius $r$ is an use-case-specific hyperparameter.

%Table~\ref{tab:Airf_sm_comp}, indicates that the training loss is similar across different GPU runs, showing that MPNN-MPI does work in the unstructured grid setting.  Here, again we see that GCN errors become worse than in the structured case~\ref{sec:darcy_flow}. We compare the errors for the configuration of a single GPU, 2 GPUs, and 4 GPUs. We see that the total loss does not vary much across all the GPUs showcasing the effectiveness of MPNN-MPI. 

\subsubsection{High-fidelity AirfRANS dataset}
For the high-fidelity mesh analysis, we employ the standardized AirfRANS dataset as detailed by ~\cite{bonnet2022airfrans}. {Our focus in this study is to show the effectiveness of DS-MPNN across various network hyperparameters rather than to achieve the best accuracy. Hence, we use a specific subset of the original AirfRANS dataset, termed as the `scarce' dataset by the authors, that consists of 180 training samples and 20 test samples.} The dataset encompasses a Reynolds number variation between 2 million and 6 million, and the angle of attack ranges from \(-5^{\circ}\) to \(15^{\circ}\). The input node attributes bear resemblance to the low-fidelity dataset~\ref{sec:low_fid} but with an expanded dimension of unit surface outward-pointing normal for the node \(v_{i}^{l={0}} \in \mathbb{R}^7\), and the edge attributes are characterized by \(e_{ij}^{(l)} \in \mathbb{R}^{10}\). The model output is denoted as \(y_{i} \in \mathbb{R}^4\). Table~\ref{tab:Airf_big_comp} presents the test error, quantified as the Root Mean Square Error (RMSE) loss, where the dataset is sampled once. These results indicate that the DS-MPNN framework maintains its efficacy even with a substantial increase in the node count in the dataset, consistently performing well across different GPU configurations. In contrast, the GCN exhibits poorer performance under similar conditions. Additionally, an experiment involving a network configuration without inter-GPU communication, but with distinct domains distributed across GPUs, resulted in divergent training outcomes which are different from~\cite{stronisch2023multi} where no communication model worked better. This underscores the critical role of communication between GPUs for maintaining model stability and performance.
Table~\ref{tab:Airf_big_comp} shows that using the DS-MPNN framework with more GPUs increases training and inference speeds. This is because each GPU deals with fewer edges and nodes than it would if the entire domain was on a single GPU, which speeds up model execution. The exploration of scalability and the associated communication overhead for DS-MPNN is comprehensively addressed in Section~\ref{sec:scale}.

\begin{table}[ht]
\centering
\caption{Comparison among GCN, S-MPNN and DS-MPNN4 for AirfRANS dataset.}

\label{tab:Airf_big_comp}
%\resizebox{\columnwidth}{!}{%
\begin{tabular}{@{}lccc@{}}
\toprule
Method & GCN & S-MPNN & DS-MPNN4 \\
\midrule
Test Error & 0.32 & 0.17 & 0.18 \\
Train Time (s/epoch) & 68 & 197 & 113 \\
Inference Time (s) & 7.05 & 17.7 & 11.7 \\
\bottomrule
\end{tabular}%
%}
\end{table}

Training cases with 3.5 million parameter models and different hyperparameters are showcased when trained on 4 GPUs for 400 epochs. On 3.5 million parameters, a single GPU has a memory overflow on Nvidia RTX6000, when trained on $s = 6000$ nodes, $n_e = 64$ sampled edges, and $h = 4$ radius hops, but as we increase the number of nodes, which is possible only through DS-MPNN, we see that the RMSE test accuracy increases in table~\ref{tab:performance_metrics_nodes}. Similarly, table~\ref{tab:performance_metrics_edges}, demonstrates the need for DS-MPNN to enable the sampling of a higher number of edges.

\begin{minipage}{0.45\textwidth}
    \centering
    \captionof{table}{Effect of increasing nodes on the performance for DS-MPNN4.} % Use \captionof{table} in minipage
    
    \label{tab:performance_metrics_nodes}
    \begin{tabular}{cccc}
        \toprule
        Nodes & Hops & Edges & RMSE Loss \\
        \midrule
        3000 & 4 & 64 & 0.27 \\
        6000 & 4 & 64 & 0.21 \\
        10000 & 4 & 64 & 0.20 \\
        \bottomrule
    \end{tabular}
\end{minipage}%
\hfill % This puts space between the minipages
\begin{minipage}{0.45\textwidth}
    \centering
    \captionof{table}{Effect of increasing edges on the performance for DS-MPNN4.}
    
    \label{tab:performance_metrics_edges}
    \begin{tabular}{cccc}
        \toprule
        Nodes & Hops & Edges & RMSE Loss \\
        \midrule
        6000 & 4 & 16 & 0.26 \\
        6000 & 4 & 32 & 0.24 \\
        6000 & 4 & 64 & 0.21 \\
        \bottomrule
    \end{tabular}
\end{minipage}

\subsection{Three-dimensional step flow dataset}
To assess the applicability of DS-MPNN for a three-dimensional problem, a canonical dataset of flow over a backward-facing step is introduced. Each flow configuration includes approximately $180,000$ nodes with varying step heights (from $0$ to $0.8$ of the channel height) and is solved using the steady Reynolds-Averaged Navier-Stokes (RANS) equations -- similar to AirfRANS but in three dimensions. The boundaries in the spanwise and flow-normal directions are walls, and the inlet maintains a constant velocity of $1$ m/s. This dataset comprises 67 training samples and 11 test samples. The neural network model trained on this dataset consists of 3.2 million parameters. The input and output dimensions mirror those of the AirfRANS dataset, with additional dimensions for the $z$--axis and $w$--velocity components. Key hyperparameters include 4 hops ($h=4$) and a sampling radius of $r=0.05$.

\begin{table}[ht]
\centering
\caption{Comparison of S-MPNN and DS-MPNN4 for the 3-D step flow dataset.}
\label{tab:3-D_step}
\begin{tabular}{@{}lccc@{}}
\toprule
Method & S-MPNN & DS-MPNN4 \\
\midrule
Test Error & 0.860 & 0.862 \\
\bottomrule
\end{tabular}
\end{table}

\begin{figure}[ht]
\centering
\includegraphics[width=\linewidth]{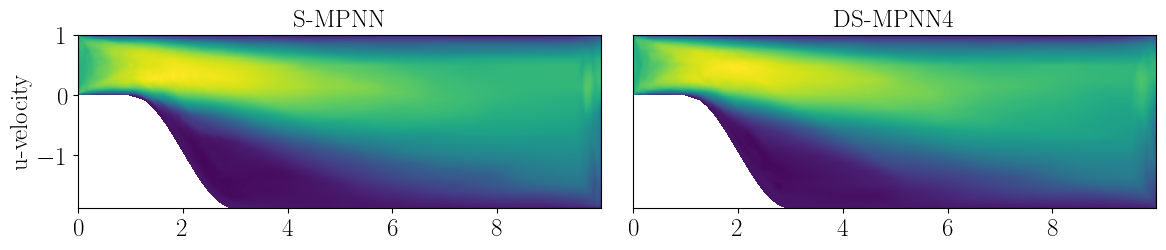}
\caption{Comparison of $x$--velocity (velocity in the direction of flow) between single GPU (S-MPNN) and four-GPU (DS-MPNN4) for a test sample of the 3-D step dataset.}
\label{fig:2d-step}
\end{figure}

Table~\ref{tab:3-D_step} and figure~\ref{fig:2d-step} illustrate that the test errors for S-MPNN and DS-MPNN4 are similar, indicating comparable performance between the single-GPU and multi-GPU implementations for this dataset. It is also important to note that the accuracy in the recirculation region of the step does not change across S-MPNN and DS-MPNN4, showing that the distributed training approach can capture the dynamically important regions.
A detailed comparison of S-MPNN and DS-MPNN4 for 3-D flow predictions is provided in the Appendix.
 
\section{Conclusions}
We address the issue of scaling edge-based graph methods to larger numbers of nodes by introducing distributed training for message-passing neural networks (MPNNs). When combined with node sampling techniques, this distributed approach allows us to scale a larger number of nodes with no or minimal loss in accuracy as compared to the single-GPU implementation while achieving a considerable decrease in training and inference times. We also show comparisons to the graph convolution networks (GCNs) and establish that edge-based methods like MPNNs outperform node-based methods like GCNs. This work opens up new avenues for the use of edge-based graph neural networks in problems of practical interest where the number of nodes can be impractically large,  $\sim O(10^5-10^6)$, for single-GPU memory limits. The limitations of DS-MPNN lie in the complexity of managing unstructured grids and adjacent elements. However, distributed graph training is advancing rapidly, with new libraries such as PyTorch Geometric version 2.5.0~\citep{fey2019fast} addressing these challenges. Future work should also involve the use of more sophisticated partitioning methods employed in engineering simulations, such as METIS.

\newpage
\bibliography{DS-MPNN.bib}
\bibliographystyle{neurips_2024}
%\bibliographystyle{plainnat}
%%%%%%%%%%%%%%%%%%%%%%%%%%%%%%%%%%%%%%%%%%%%%%%%%%%%%%%%%%%%%%%%%%%%%%%%%%%%%%%
%%%%%%%%%%%%%%%%%%%%%%%%%%%%%%%%%%%%%%%%%%%%%%%%%%%%%%%%%%%%%%%%%%%%%%%%%%%%%%%
% APPENDIX
%%%%%%%%%%%%%%%%%%%%%%%%%%%%%%%%%%%%%%%%%%%%%%%%%%%%%%%%%%%%%%%%%%%%%%%%%%%%%%%
%%%%%%%%%%%%%%%%%%%%%%%%%%%%%%%%%%%%%%%%%%%%%%%%%%%%%%%%%%%%%%%%%%%%%%%%%%%%%%%
\newpage
\appendix
\section{Ablation Studies on Scalability and Communication Overhead}
\label{sec:scale}
\begin{figure}[htbp]
    \centering
    
    \begin{minipage}{0.48\textwidth}
        \centering
        \includegraphics[width=\linewidth]{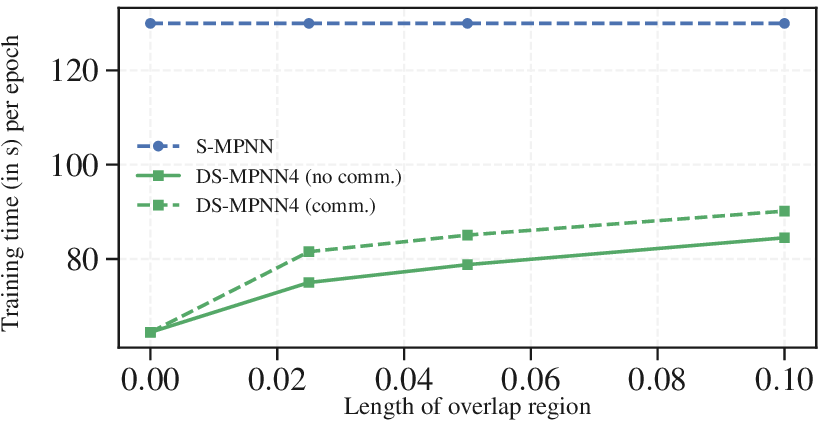}
         % Adjust the vertical space between the images as needed
        
        \includegraphics[width=\linewidth]{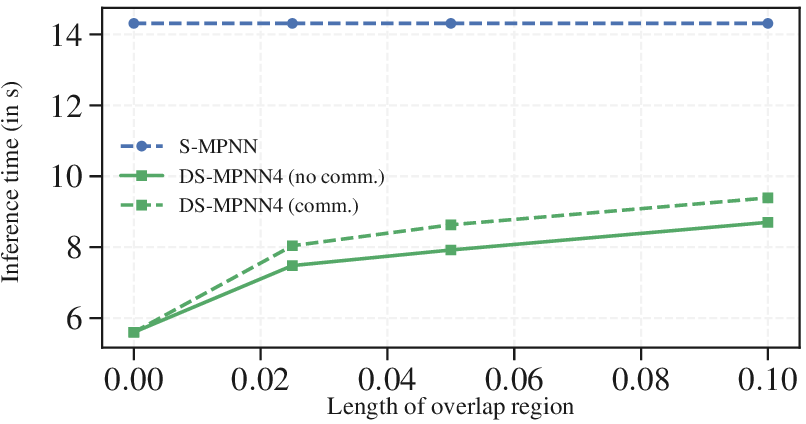}
        \caption{Training time per epoch and inference time as a function of the length of the overlap region ($l$).}
        \label{fig:overlap-train_inference-time}
    \end{minipage}\hfill
    \begin{minipage}{0.48\textwidth}
        \centering
        \includegraphics[width=\linewidth]{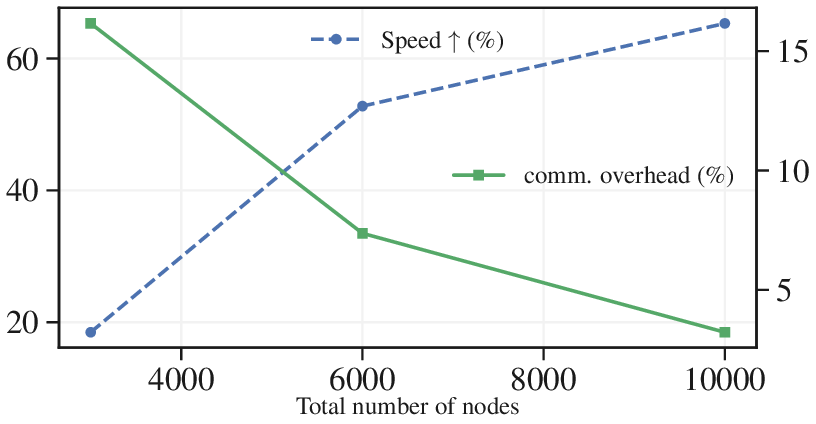}
        \caption{Performance of DS-MPNN4 against S-MPNN with node increase.}
        \label{fig:speed_comm_overhead}
        \vspace{0.5cm} % Adjust the vertical space between the images as needed
        \includegraphics[width=\linewidth]{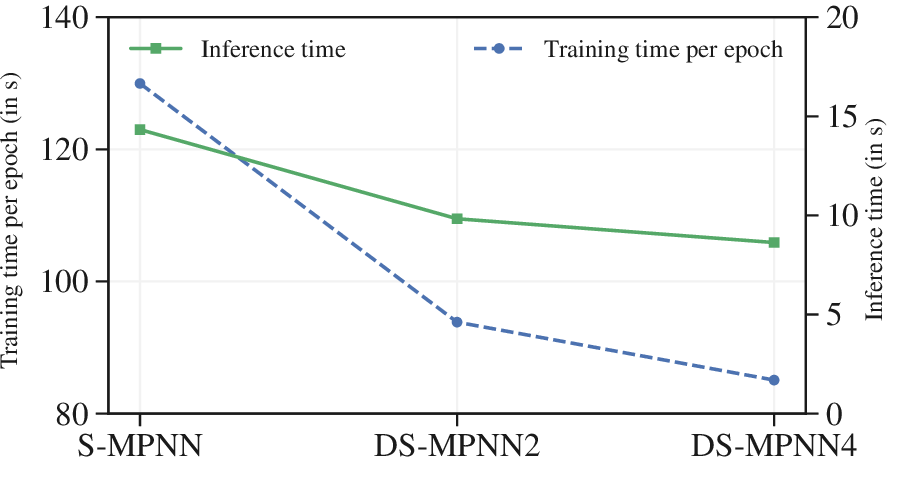}
        \caption{Training and inference time as a function of the number of distributed domains.}
        \label{fig:mpnn_performance}
    \end{minipage}
    
\end{figure}

{The subsequent ablation experiments were conducted to evaluate the scalability and communication burden of DS-MPNN4 in comparison with the standard single GPU framework, S-MPNN. These studies were carried out using the `scarce' high-fidelity AirfRANS dataset, which consists of 180 training samples. For these experiments, the kernel radius and the number of nodes were predetermined at $r=0.05$ and $6000$, respectively. Unless specifically mentioned, the baseline overlap length is set to the kernel radius. Figure~\ref{fig:overlap-train_inference-time}, clearly demonstrate the associated cost of communication for DS-MPNN4. The training and inference costs for communication are higher compared to scenarios with no communication across various non-zero lengths of the overlap region in the 4 sub-domains. Nevertheless, despite the communication overhead, DS-MPNN4 is consistently better than S-MPNN in terms of training and inference duration.}
We also increase the sampled node count $s$ to ascertain the impact of scalability and communication overhead. This increase in node samples becomes imperative for larger numerical grids. Figure~\ref{fig:speed_comm_overhead} shows the increase in the training speed of DS-MPNN4 compared to S-MPNN. The increase in speed is attributed to the reduction in the time required for graph kernel $\mathcal{G}$ generation within DS-MPNN4, as a consequence of decreased number of edge formations required across each GPU domain. Correspondingly, the percentage of training time spent on communication decreases as the node count increases, because, while the aggregate data for communication grows with the node count, the substantial bandwidth of GPUs maintains a constant communication time.
\review{Moreover, scalability appears in figure~\ref{fig:mpnn_performance} that shows a decrease in training and inference time with an increasing number of domains from one (S-MPNN) to four (DS-MPNN4) GPU setup. This is consistent with our previous observations.}
\clearpage
\section{Darcy Flow Equations}

2-D Darcy flow equation on the unit box is a second-order elliptic PDE given as:

\begin{equation}
    - \nabla (a(x)\nabla u(x)) = f(x) \quad \forall  x \in (0,1)^2 
\end{equation}

\begin{equation}
    u(x) = 0  \quad \forall  x \in \partial(0,1)^2.
\end{equation}

Here, $a$ and $f$ are the spatially varying diffusion coefficients and the forcing field, respectively. $u$ is the solution field on the 2-D domain. This example parallels the approach used in the Graph Neural Operator (GNO) as discussed by ~\cite{li2020neural}, adapted for a single GPU setup with a notable distinction: we update the edge attributes following each radius hop or message-passing step, and both edges and nodes are subjected to sampling. The experiment investigates the mapping \( a \rightarrow u \).

\section{Darcy Flow Visualizations}
\begin{figure}[ht]
\centering
\includegraphics[width=0.6\linewidth]{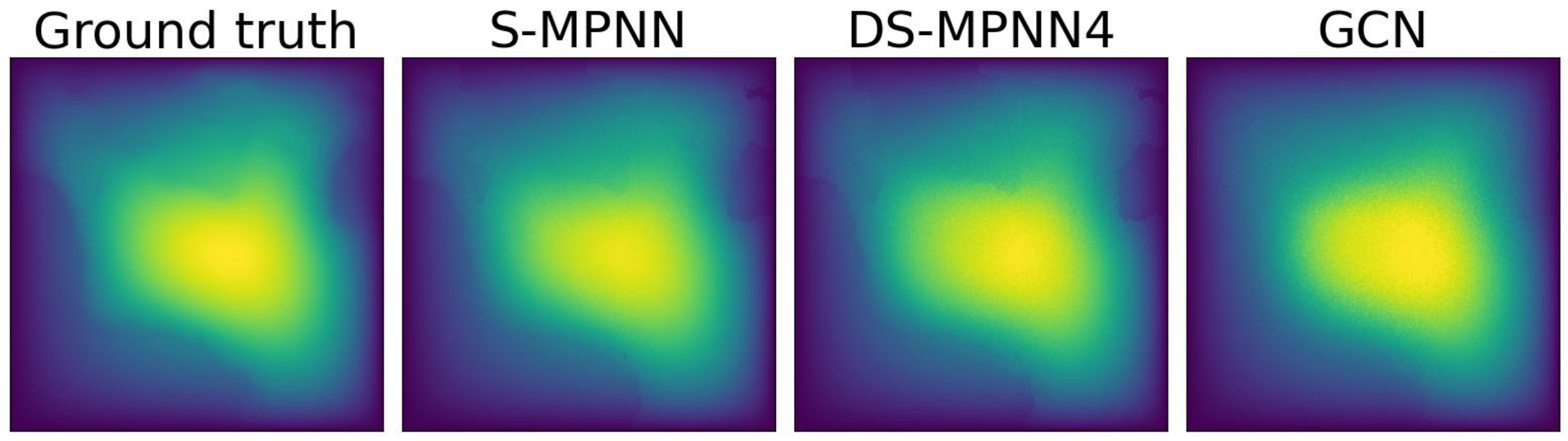}
\caption{Comparison of model performance between single GPU (S-MPNN) -- \review{RMSE=$5.3\times 10^{-6}$}, four-GPU (DS-MPNN4) -- \review{RMSE=$4.71\times 10^{-6}$}, and GCN -- \review{RMSE=$4\times 10^{-3}$}, for a test sample of Darcy flow dataset.}
\label{fig:Darcy}
\end{figure}

Figure~\ref{fig:Darcy} illustrates that both DS-MPNN and S-MPNN configurations provide similar predictions for the Darcy flow dataset. In contrast, the GCN model significantly under-performs, as it excessively smooths the solution and fails to capture the roughness at the boundaries accurately.
%%%%%%%%%%% figure 1 %%%%%%%%%%%%%%%%%%%%

\section{AirfRANS equations}

The AirfRANS dataset corresponds to the solution of incompressible steady-state Reynolds Averaged Navier Stokes (RANS) equations in 2-D,  given by

\begin{equation}
\mathbf{U}.\nabla \mathbf{U} = -\frac{1}{\rho}\nabla P + (\nu + \nu_t)\nabla^2 \mathbf{U},
\end{equation}
    
\begin{equation}
\nabla.\mathbf{U} = 0.
\end{equation}

Here, $\mathbf{U}=[U_x, U_y]$ represents the mean 2-D velocity components, and $P$ denotes the mean pressure. The variable $\nu_t$ signifies the kinematic turbulent viscosity, which is spatially varying, while $\nu$ denotes the constant kinematic viscosity. In this dataset, the learning process involves mapping the airfoil shape and angle of attack to $[U_x, U_y, P, \nu_t]$, as further elaborated in the subsequent sections.

\section{Three-dimensional step flow visualizations}

\begin{figure}[ht]
\centering
\includegraphics[width=0.48\linewidth]{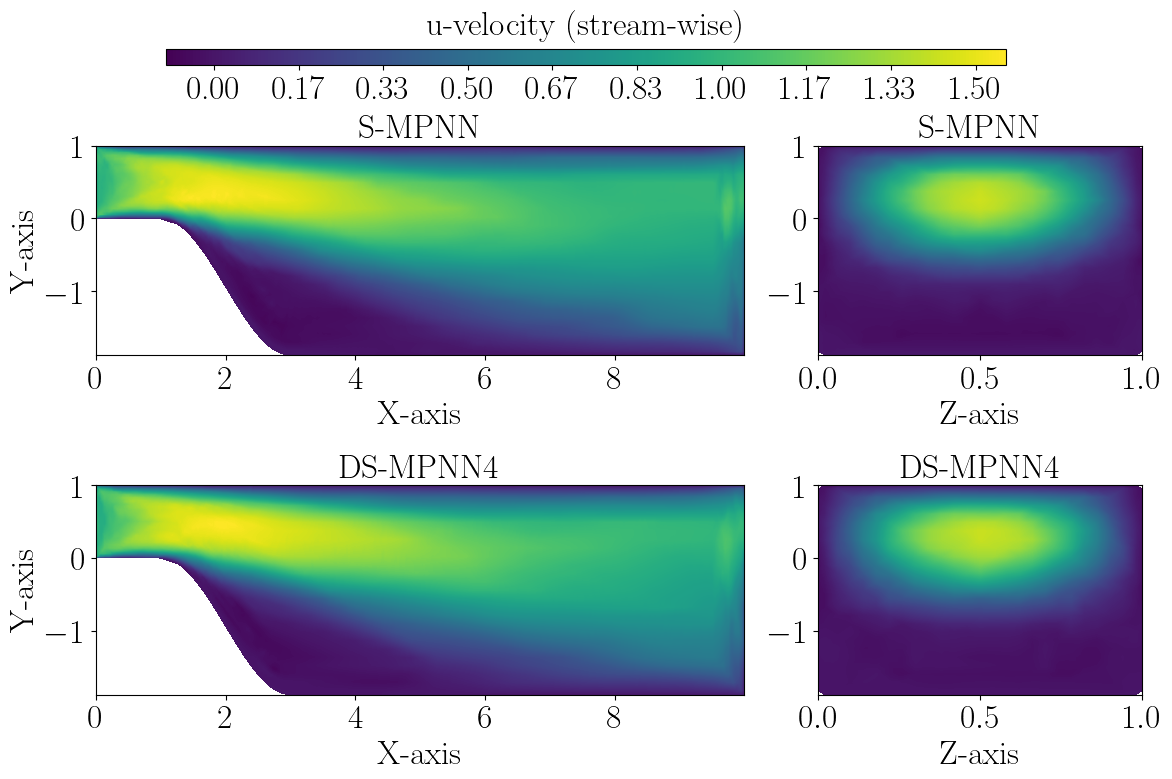}
\includegraphics[width=0.48\linewidth]{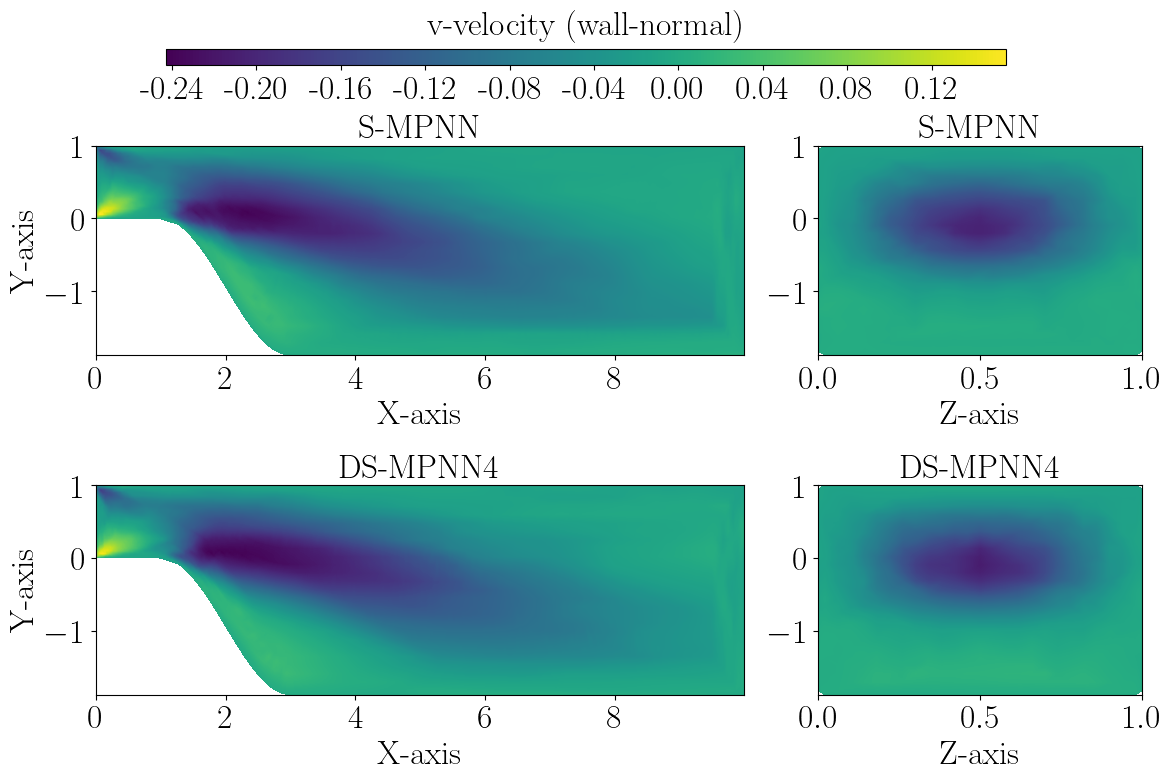}
\caption{Streamwise \( u \)-velocity prediction (left) and wall-normal \( v \)-velocity (right). The xy-slice is at \( z = 0.5 \) and the yz-slice is at \( x = 3.5 \). The top row depicts S-MPNN predictions, while the bottom row shows DS-MPNN4 predictions.}
\label{fig:3_d_uvel}
\end{figure}

\begin{figure}[ht]
\centering
\includegraphics[width=0.48\linewidth]{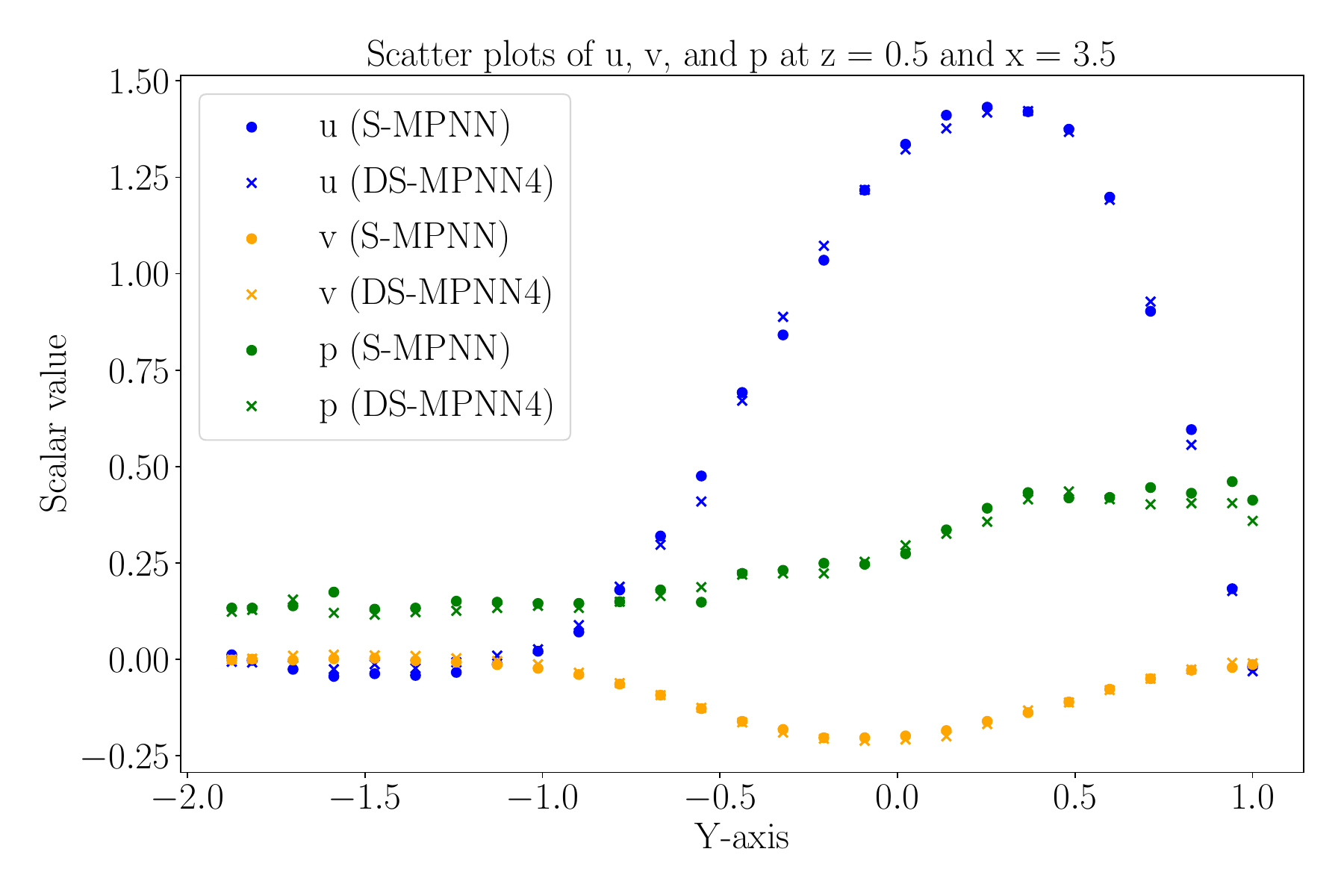}
\caption{Comparison of S-MPNN and DS-MPNN4 at \( x = 3.5 \) and \( z = 0.5 \).}
\label{fig:3_d_uvel_Scatter}
\end{figure}

The predictions for S-MPNN and DS-MPNN4 are further illustrated in Figure~\ref{fig:3_d_uvel}, where two slices are presented: one at \( z = 0.5 \) in the xy-plane, and the other at \( x = 3.5 \), which lies within the recirculation region. The results demonstrate the accuracy and consistency of both models. The scatter plot in Figure~\ref{fig:3_d_uvel_Scatter} further corroborates this by comparing the predictions at \( z = 0.5 \) and \( x = 3.5 \), indicating a high degree of agreement in the predictions of two models.

\newpage
\section{Training Algorithm}
\begin{algorithm}
    \caption{Training of DS-MPNN.}
    \label{alg: Train MPNN-MPI}
    \KwIn{ Node encoder: $\mathcal{N}_e$, Node decoder: $\mathcal{N}_d$, Graph kernel: $\mathcal{K}_\phi$; Domain: $\Omega$, Sampled-domain: $\Omega_{s}$, Sub-domain: $\Omega_{r}$;  grid coordinates and initial conditions: $v^{l={0}}$., Target values: $Y$, No of training samples: $N_{train}$, Number of epochs: $E$, Number of distributed domains: $n_{proc}$.}
    \For{$j$ = $1$ \KwTo $N_{train}$}{
    $\Omega_{s} \leftarrow rand(\Omega)$ \Comment{Sample nodes from domain $\Omega$} \\
    $\Omega_{r} \leftarrow \Omega_{s}$ \Comment{Decompose domain based on available gpus}\\
    $v^{l={0}} \leftarrow \Omega_{r}$ \Comment{Grid coordinates and initial values to create nodes}\\
    $Y \leftarrow \Omega_{r}$ \Comment{Get targets in the sub-domain}\\
    $edge_{I} \leftarrow edgeindexcreator(\Omega_{r},r)$ \Comment{Create edge index  for kernel radius r}\\
    $edge_{i} \leftarrow rand(edge_{I})$ \Comment{Sample edge index}\\
    $e^{l} \leftarrow (dv^{l={0}})$ \Comment{Create edge attributes ($dv = v_{i}-v_{j})$}\\
    $trainloader \leftarrow tuple(v^{l={0}},Y,edge_{i},e^{l})$
    }
     \For{$\textrm{epoch} = 1$ \KwTo $E$} 
     {\Comment{This loop run across $n_{proc}$ GPUs} \\
    $v^{l={0}},Y,edge_{i},e^{l} \leftarrow trainloader$\\
    $i_{b} \leftarrow getindex(v^{l={0}})$ \Comment{Tracking index $i_b$ for unstructured grids} \\
    \For{$k$ = $1$ \KwTo $h$}
    {   \If{k = 1}{
        ${v^{l}_{L}}\leftarrow \mathcal{N}_{e}\left({v^{l={0}}},{w}\right)$ \Comment{Node encoding}}
        $v_{res} \leftarrow v^{l}_{L}$
        $v^{l}_{L} \leftarrow \mathcal{K}_{\phi} \left({v^{l}_{L}},edge_i,e^{l},{w}\right)$  \Comment{Graph convolution}\\
        $v^{l}_{L} \leftarrow v^{l}_{L} + v_{res}$
        $v_{i}^{l}\leftarrow \mathcal{N}_{d}\left(v^{l}_{L},{w}\right)$ \Comment{Node decoding}\\
        $v^{l}_{L} \leftarrow Comm(i_b,\Omega$,$v^{l}_{L}$) \Comment{Communicate node values in latent space} \\
        $v^{l} \leftarrow Comm(i_b,\Omega,v^{l})$ \Comment{Communicate output values}  \\
       $e^{l} \leftarrow dv^{l}$ \Comment{Update edge attributes with new values}} 
        \vspace{2 mm}
         $\mathcal{L}_{local}$ =  $\textrm{MSE}\left({{{v^{l}}_{i_{(int)}}},{Y_{i_{(int)}}}}\right)$ \Comment{MSE on interior points}\\
         $\mathcal{L} =  \sum_{i=1}^{n_{proc}}\mathcal{L}_{local} $ \Comment{Getting loss from all procs}\\
        $\nabla {w_{local}} \leftarrow  \textrm{Backprop}(\mathcal{L})$ \Comment{Get local gradients}\\
        $\nabla {w}$ =  $\sum_{i=1}^{n_{proc}}\nabla {w_{local}}$ \Comment{Sum local gradients}\\
        ${w} \leftarrow {w} - \eta \nabla {w}$ \Comment{Update parameters}\\
    }
    \KwOut{Trained DS-MPNN model $\mathcal{N}(\mathcal{N}e,\mathcal{N}d, \mathcal{K}\phi)$}
\end{algorithm}

\clearpage

\end{document}